\documentclass[sigconf]{acmart}
\usepackage[capitalise]{cleveref}
\usepackage{tabularx}
\usepackage{makecell}
\usepackage{subcaption}
\AtBeginDocument{%
  }

\copyrightyear{2024}
\acmYear{2024}
\setcopyright{rightsretained}
\acmConference[RecSys '24]{18th ACM Conference on Recommender Systems}{October 14--18, 2024}{Bari, Italy}
\acmBooktitle{18th ACM Conference on Recommender Systems (RecSys '24), October 14--18, 2024, Bari, Italy}\acmDOI{10.1145/3640457.3691711}
\acmISBN{979-8-4007-0505-2/24/10}

\begin{document}

\title{Understanding Fairness in Recommender Systems:\\ A Healthcare Perspective}

\author{Veronica Kecki}
\email{verokecki@gmail.com}
\orcid{0000-0001-6229-7792}
\affiliation{%
  \institution{University of Gothenburg}
  \city{Gothenburg}
  \country{Sweden}
}
\author{Alan Said}
\email{alan@gu.se}
\orcid{0000-0002-2929-0529}
\affiliation{%
  \institution{University of Gothenburg}
  \city{Gothenburg}
  \country{Sweden}
}

\renewcommand{\shortauthors}{Kecki and Said}

\begin{abstract}
Fairness in AI-driven decision-making systems has become a critical concern, especially when these systems directly affect human lives. This paper explores the public's comprehension of fairness in healthcare recommendations. We conducted a survey where participants selected from four fairness metrics -- Demographic Parity, Equal Accuracy, Equalized Odds, and Positive Predictive Value -- across different healthcare scenarios to assess their understanding of these concepts. Our findings reveal that fairness is a complex and often misunderstood concept, with a generally low level of public understanding regarding fairness metrics in recommender systems. This study highlights the need for enhanced information and education on algorithmic fairness to support informed decision-making in using these systems. Furthermore, the results suggest that a one-size-fits-all approach to fairness may be insufficient, pointing to the importance of context-sensitive designs in developing equitable AI systems.
\end{abstract}



\maketitle

\section{Introduction}
The rapid advancements in AI have significantly transformed various sectors, including healthcare, finance, and education. These systems have the potential to enhance decision-making processes, improve efficiency, and provide personalized recommendations. However, integrating AI into critical decision-making roles raises significant ethical concerns. Algorithmic fairness, in particular, has become a major issue, especially when these systems affect individuals' lives and well-being \cite{schafer_towards_2017,hauptmann_research_2022}.

In healthcare, the stakes of algorithmic decisions are exceptionally high, as they can influence patient outcomes, resource allocation, and overall quality of care \cite{wu_development_2020}. Ensuring that these systems operate fairly and do not perpetuate biases is crucial. Various fairness metrics have been proposed to address these concerns, including Demographic Parity, Equal Accuracy, Equalized Odds, and Positive Predictive Value. Each metric offers a different approach to fairness, highlighting the complexity and contextual nature of defining and achieving fairness in AI systems. The computer science literature has amassed more than twenty different notions of fairness, making it infeasible to create a ``catch-all'' metric \cite{abu_elyounes_contextual_2019}.

In recommendation and decision-making, any machine learning model will inevitably alleviate one condition while simultaneously deteriorating another \cite{hiller_fairness_2020}, a phenomenon known as competing notions of fairness, or ``trade-offs'' \cite{samuel_why_2022}. It is impossible to definitively state that one fairness metric is ``fairer'' than another in a general sense. A metric may be fairer in one aspect and less fair in another. For instance, a model could be optimized for either equality or accuracy, but it can be challenging to determine which is more appropriate in a given context. This raises the question: who decides what is fair?

This study investigates the public’s understanding and perception of fairness in healthcare recommender systems. By conducting a survey in which participants chose among commonly used fairness metrics in various healthcare scenarios, we aimed to uncover how people perceive and understand fairness and how these perceptions change based on the context of the decision-making situation. Our study reveals a low level of public understanding of fairness and significant influence from the specific context of the decision. Additionally, there is generally a lack of consensus on what fairness means in different scenarios.

We argue that improving knowledge about algorithmic fairness is essential for making informed choices when using AI-driven decision support systems.

\section{Related Work}
Fairness in algorithmic decision-making is crucial, particularly in recommender systems that impact critical areas like healthcare \cite{schafer_towards_2017}. Various fairness metrics, such as Demographic Parity, Equalized Odds, and Positive Predictive Value, address different aspects of fairness. For instance, Demographic Parity focuses on equal representation, while Equalized Odds ensures equal error rates across groups. These metrics highlight the trade-offs between fairness and accuracy, especially in high-stakes scenarios like healthcare.

A central aspect in the discussion of algorithmic fairness which helps to understand the difference between different metrics and the involved trade-offs is individual vs. group fairness, with accuracy typically representing the former and parity addressing the latter. A study by \citet{schlicker_what_2021}, for instance, contrasted people’s perceptions of fairness between human and automated agents pointing to a higher degree of consistency in automated decision-making. Similarly, \citet{kochling_highly_2021} also found that algorithms tend to carry out more accurate decisions than human agents; at the same time, the authors of this study highlighted the fact that despite the high degree of accuracy, there is simultaneously the tendency to perpetuate biases. There are quite a few studies which emphasize the need for optimizing the algorithms in such a way that protect vulnerable populations (\citet{rajkomar_ensuring_2018}, for instance, propose concrete machine learning solutions to that effect), but it should be kept in mind that such strategies always come at the cost of accuracy on an individual level. In other words, the inherent conflict of algorithmic decision-making is that automated decisions are either likely to be accurate for individuals while enforcing biases on a group level or if group biases are mitigated, it will lower the degree of accuracy on an individual level.

While the literature on people’s fairness preferences in general socio-economic contexts is quite extensive, it still remains rather scarce when it comes to either the healthcare context or more specifically algorithmic decision-making. The former can be exemplified by the study by \citet{li2019} which found that public preference tends to lean towards accuracy over demographic parity when patient outcomes are at stake. As for the latter, \citet{srivastava_mathematical_2019} noted a general preference for demographic parity, but with a shift towards accuracy in high-stakes situations. This suggests that fairness is highly contextual and should align with societal values to ensure equitable outcomes. 

Public perception significantly influences the acceptance of recommendation systems which underscores the need for transparency and public involvement. While fairness has been extensively researched in recommender systems, e.g., \citet{li_tutorial_2021} focusing on foundations and algorithms for fairness, \citet{elahi_beyond_2021} investigating evaluation from a beyond algorithm perspective, and \citet{wu_fairness_2023} studying evaluation approaches and assurance strategies for fairness, the concept of fairness understanding among the end-users of recommender and decision support systems remains understudied \citet{wang_survey_2023}. It is also worth noting that some of the studies that have explored people's fairness preferences in the context of automated decision-making and were able to identify a clear preference were based on case studies where the stakes were not strongly contrasted (see \citet{saxena_how_2019}).
\begin{table*}[t]
\caption{The answer options available (A through E) in the study, including the metric and descriptions, as presented to the study participants in the high-stake scenario.}
\label{tab:algfair}
\begin{tabularx}{\linewidth}{llX}

\toprule
\textbf{Option} & \textbf{Metric} & \textbf{Description} \\
\hline
A & Demographic Parity & The same percentage of patients from both groups, regardless of whether they actually need a transfer or not (i.e., if they are qualified), get transferred to the ICU—but the algorithm may not be equally accurate across the two groups (i.e., it may happen that the accuracy is much lower for either of the groups)\\
\hline
B & Equal Accuracy & The model is equally accurate for both groups (i.e., same percentage of True Negatives + True Positives)—but it may be so that a smaller portion of patients from either of the groups overall ends up getting placed at the ICU (in this case, irrespective of being qualified)\\
\hline
C & Equalized Odds & The same percentage of patients actually requiring a transfer (i.e., those who qualify) end up getting one across both groups—but it could happen that a much smaller overall portion of patients from either group get the allocation\\
\hline
D & Positive Predicted Value & Out of all the patients who end up getting a spot at the ICU (i.e. True Positive + False Positive), the same portion is identified correctly by the algorithm across both groups— but that could include a much lower portion of patients from one of the groups overall, as well as it may imply lower accuracy for one of the groups\\
\hline
E & N/A & I do not understand the options\\
\bottomrule
\end{tabularx}
\end{table*}

\section{Study Design}
\label{sec:method}
We conducted a survey where participants were tasked with choosing between four different algorithmic fairness metrics in a healthcare setting.

The survey comprised three parts: an introductory section with background information, two main questions, and several supporting questions. This description primarily focuses on the two main questions. It should be noted that we accounted for the possibility that respondents might have little to no knowledge of the subject. Therefore, the descriptions and questions gradually introduced the topic over the course of the survey.

The survey began with a short background introducing the issue of competing notions of fairness and briefly explaining the options presented in the two main questions.

The specific task was presented to the participants as follows:\\
\indent \emph{You will be presented with two different scenarios taking place in the healthcare context. An AI system needs to make a decision resulting in the allocation of resources across 2 equally-sized groups: a high-income patient group, and a low-income one. The choice needs to be made between 4 different algorithmic fairness metrics which prioritize different conditions.}\\

Following this description, the participants were presented with two scenarios, each correlating to a hypothetical medical situation, i.e., one high-stake and one low-stake.
Respondents were asked to choose between four different options (each optimizing a different fairness metric) that they felt achieved the best overall fairness in these two scenarios: \\

\indent\textbf{Scenario 1 (high-stake):} Imagine that a hospital is using an AI-based monitoring system to warn the rapid response team about patients at a high risk for deterioration, requiring their transfer to an intensive care unit within 6 hours.\\ 
\indent\textbf{Scenario 2 (low-stake):} Imagine that a hospital is using an AI tool that allows to identify patients who are likely to develop vitamin deficiencies from the existing patient health data, and may subsequently recommend such patients to get a blood test.\\

The high-stake condition in Scenario 1 was identical to \cite{rajkomar_ensuring_2018}, whereas the condition in Scenario 2 was adapted for the survey and converted into a low-stake condition. The fairness options presented under both scenarios were kept identical with the aim of contrasting these two conditions to be able to see \emph{1)} if there is any difference at all between participants' preferences depending on the context (i.e., gravity of situation), and \emph{2)} which factors it could possibly be attributed to if detected. Furthermore, the study also contrasted between two different groups of patients: a high-income patient group and a low-income one, in order to compare fairness between both groups. For the sake of simplicity, the study presumes that the groups are equally-sized. 

The four algorithmic fairness metrics chosen for the study were \emph{Demographic Parity}, \emph{Equal Accuracy}, \emph{Equal Odds}, and \emph{Positive Predictive Value} \cite{rajkomar_ensuring_2018}. Each of these metrics alleviates at least one condition and deteriorates one other. Demographic Parity is achieved if two populations are equally represented in the outcome, independent of the size of the populations - in this scenario, demographic parity is achieved if patients from the high-income and low-income groups are equally represented in the outcome. Equal Accuracy, also referred to as accuracy parity, requires the model to perform equivalently (in terms of, e.g., prediction accuracy) within the populations in order to attain equal accuracy. Equalized odds refers to the notion of two equivalently qualified data points in two populations having the same probability of being selected, independent of population sizes. Finally, Positive predictive value, or precision, specifies the fraction of predicted values being correct. The specific formulations used in the high-stake scenario in survey are shown in \cref{tab:algfair}. Similar descriptions were also presented for the low-stake scenario, although worded so as to reflect the low-stake of that scenario.

The difference between the response options for both scenarios is that the first scenario involves a high-stake condition (i.e., life/death situation), and the second question involves a low-stake condition (i.e., ``nice-to-have'' situation). 

The rationale behind the questions corresponding to different conditions (high-stake/low-stake) is to see if a difference in preference for a particular option can be detected depending on the context. As seen in \cref{tab:algfair}, an additional option (E) was included, for the survey participants to be able to indicate that they do not fully understand the four model options (whether individually or the difference between them).

Responses were collected through an online survey (Google Forms). 
The link to the survey was circulated on social media from the personal and professional profiles of a number of people and organizations within the higher education sector, reaching approximately 10, 000 followers combined.  
The participants had no knowledge about the specific algorithmic models prior to the study, although many expressed familiarity with the general subject matter.

\section{Data}

A total of 131 survey responses were collected, representing a diverse cross-section of the population. Slightly more than half of the respondents identified as male, slightly fewer than half as female, and a small percentage as non-binary or preferred not to say. The respondents' ages ranged from 19 to 59 years old, with an average age of 32 and a median age of 30. Over 60\% of respondents worked full-time, while nearly 40\% were students, indicating a broad mix of professional and educational backgrounds.

The majority of respondents (76\%) were from Sweden, the country of the paper's origin, 19\% were from the European Union, and the remaining 5\% were from other countries. This geographic distribution reflects the channels through which the survey was distributed, primarily targeting individuals in the higher education sector. The diversity in age, gender, and geographic location among respondents provides a context for interpreting the survey results, particularly in understanding the varying perceptions of fairness in healthcare recommendation systems.

The survey's design and distribution aimed to capture a wide range of perspectives, which is critical for exploring public comprehension of fairness metrics across different healthcare scenarios.

\begin{figure*}[t]
    \centering
    \begin{subfigure}[b]{0.45\linewidth}
        \centering
        \includegraphics[trim={0 5cm 0 5cm},clip,width=\linewidth]{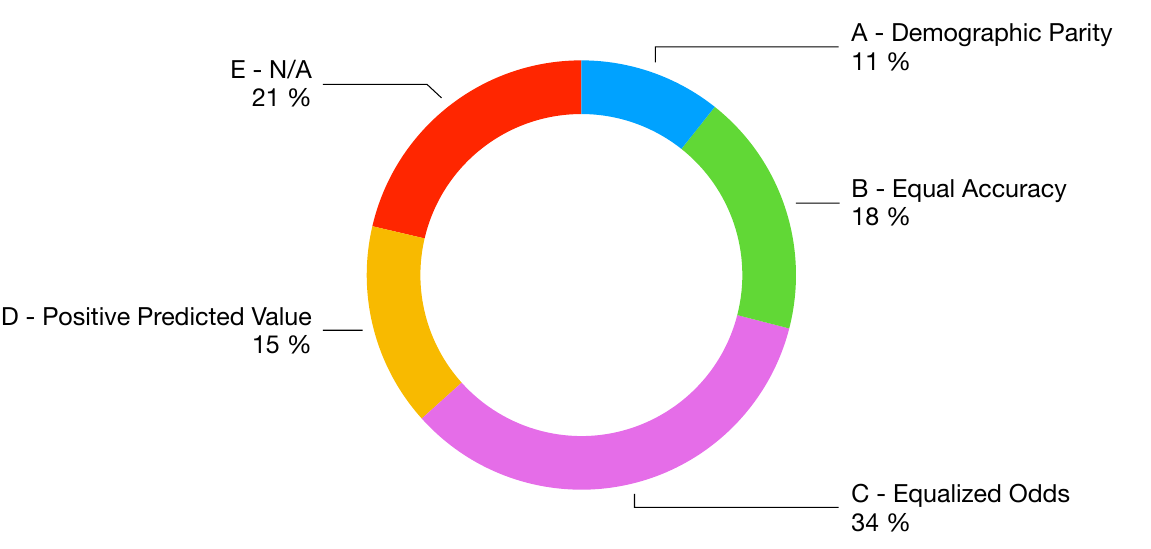}
        \caption{Scenario 1}
        \Description{A pie chart containing the answer percentage for each answer option for Scenario 1}
        \label{fig:scenario1}
    \end{subfigure}
    \hfill
    \begin{subfigure}[b]{0.45\linewidth}
        \centering
        \includegraphics[trim={0 5cm 0 5cm},clip,width=\linewidth]{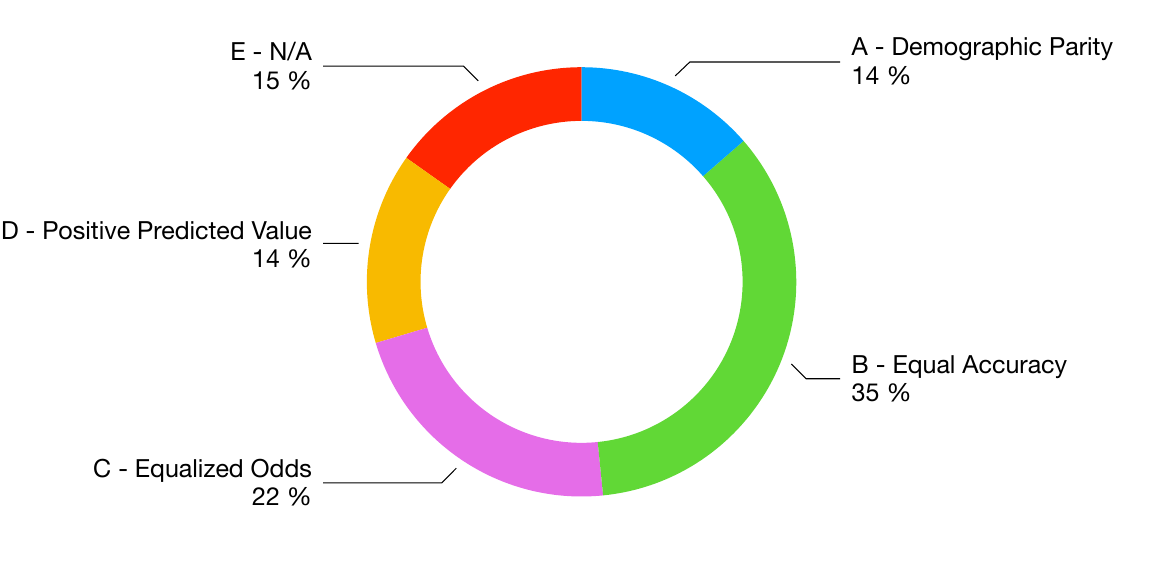}
        \caption{Scenario 2}
        \Description{A pie chart containing the answer percentage for each answer option for Scenario 2}
        \label{fig:scenario2}
    \end{subfigure}
    \caption{The percentages of respondents' preferred fairness metrics for Scenarios 1 and 2}
    \label{fig:scenarios}
\end{figure*}

\section{Results and Analysis}
\label{sec:results}
The results of the user study are presented in \cref{fig:scenario1} and \cref{fig:scenario2} for Scenario 1 and Scenario 2 respectively (cf. \cref{sec:method}). 

\cref{fig:scenario1} and \cref{fig:scenario2} show an overview of participants' responses to Scenario 1 and 2. Each option (i.e., answers A through E as presented in \cref{tab:algfair}) is represented. There is a clear preference for specific algorithmic fairness metrics in each case: Option C (Equalized Odds) was preferred for Scenario 1 (34\% of responses), and Option B (Equal Accuracy) for Scenario 2 (35\% of responses). In other words, the most common response among participants was the Equal Odds metric in a high-stake scenario and the Equal Accuracy metric in a low-stake one.

It is important to note that only one of the four metrics used in this study can be classified as an ``equality metric,'' namely, Option A, Demographic Parity. The remaining metrics -- Equal Accuracy, Equal Odds, and Positive Predictive Parity -- are variations of ``equity'' metrics \cite{abu_elyounes_contextual_2019}. These metrics prioritize accuracy, but do so in different ways.

The results indicate a trend of participants preferring accuracy over equality, regardless of the scenario's stakes. This preference suggests that when it comes to health, accuracy takes precedence, even in less critical situations. This finding aligns with previous studies, such as \citet{li2019}, which concluded that individuals prefer accuracy when lives are at stake, as opposed to equality when monetary resources are involved, and \citet{srivastava_mathematical_2019} which showed a tendency to prefer the Demographic Parity model in most contexts except for healthcare where accuracy is preferred. 

While accuracy is favored in both scenarios, the specific preferences for Equal Accuracy and Equalized Odds reflect nuanced differences in how participants perceive fairness in healthcare contexts. Equalized Odds emphasizes ensuring that individuals who genuinely need assistance receive it, making it more appropriate for high-stake scenarios where the consequences of misallocation are severe. Conversely, Equal Accuracy focuses on maintaining accuracy across all groups, which might be considered more suitable in low-stake situations where resource efficiency is important. This distinction could explain the different preferences observed between the two scenarios.

Additionally, the emphasis on accuracy in healthcare, even in low-stake scenarios, may be influenced by the intrinsic value placed on health outcomes. This prioritization of accuracy suggests that participants perceive healthcare decisions as inherently high-stakes, where the potential impact on individuals' well-being outweighs considerations of equality. The consistent preference for accuracy, regardless of the stakes, highlights the complexity of defining fairness in contexts where human health is involved.

\section{Discussion}
The results of this study reveal a certain divergence in public perceptions of fairness within healthcare recommender systems. Even when presented with scenarios that could be perceived as having objectively fair alternatives, participants displayed different preferences across several fairness metrics they were presented with. This divergence underscores a fundamental challenge in designing recommender systems that are perceived as fair by all users.


While approximately one-third of respondents favored Equalized Odds for high-stakes scenarios, whereas a similar fraction preferred Equal Accuracy for low-stakes scenarios, the remaining participants expressed varied opinions in both scenarios. This lack of consensus suggests that subjective interpretations of fairness can vary widely, even when certain metrics may appear to offer clear solutions. Factors contributing to these variations include individual values, personal experiences, and differing understandings of the implications of each of the fairness metrics. 

One notable observation is that the preference for Equal Accuracy and Equalized Odds highlights a tension between prioritizing overall accuracy and ensuring fairness in outcomes. Equal Accuracy aims to provide consistent accuracy across different groups, while Equalized Odds focuses on equalizing the chances of receiving a positive outcome for all qualified individuals. These metrics, while both emphasizing aspects of fairness, address it from different angles, illustrating the complexity of balancing fairness in practice. The disagreement among participants suggests that no single metric can be universally accepted as the most fair, as individuals weigh the importance of different fairness aspects differently.

This disparity in perceptions also points to a broader issue in the deployment of recommender systems: the communication and understanding of algorithmic decisions. The study revealed that a significant portion of participants felt unsure about the fairness metrics, as indicated by the notable selection of the ``N/A'' option. This highlights a critical need for clear and accessible explanations of how algorithms function and the trade-offs they entail. Users should understand not only the mechanics of the algorithms but also the ethical considerations involved in choosing one fairness metric over another.

Furthermore, the findings suggest that fairness is inherently subjective and context-dependent. For instance, in high-stakes healthcare scenarios, the focus on accuracy may be driven by the critical need to correctly identify patients requiring urgent care. In contrast, in low-stakes scenarios, participants may prioritize fairness in terms of equitable access to resources or recommendations, even if it means accepting some level of inaccuracy. This indicates that public perception of what constitutes fairness can shift based on the stakes involved, further complicating the task of developing universally fair systems. This indicates that public perception of what constitutes fairness can shift based on the stakes involved which also points to unfeasibility of developing universally fair systems. 

The apparent lack of a universally accepted definition of fairness in these contexts poses a challenge for developers and policymakers. It calls for a more nuanced approach to designing recommender systems, one that considers the diverse perspectives of users and strives to accommodate them. This might involve implementing adaptive systems capable of adjusting fairness metrics based on real-time context or user preferences. Additionally, engaging users in the design process through participatory methods could help align system outcomes with user expectations, thereby enhancing the perceived fairness and acceptance of the system.

This study underscores the complexity of achieving consensus on fairness in healthcare recommender systems. The varied range of opinions among participants reflects the diverse nature of fairness perceptions, which are influenced by individual and contextual factors. Addressing these differences requires thoughtful consideration of communication strategies, system design, and user engagement to create recommender systems that are both effective and perceived as fair by a broad user base.

It is worth mentioning that there remain certain practical challenges in realizing that ambition. Improving the general knowledge on the subject among the general public required to ensure proper understanding of the presented questions through educational curricula or information campaigns can take years, and until then, quite extensive background information is going to be necessary to be included into every study of such kind. That, in turn, adds to the cognitive load of participants and may affect participation rates and quality of results. It is also difficult to ensure genuine understanding of the presented options by participants. Interactive forms of data collection such as interviews or focus groups are much more likely to provide higher quality results but will fail to achieve a scale needed to be representative of the overall public opinion. Surveys, on the other hand, have the capacity to achieve that scale but offer lower guarantees of proper understanding of the questions.

\section{Conclusions}

This study highlights the complexities and challenges inherent in achieving fairness in healthcare recommendation systems. The diverse preferences observed among participants for different fairness metrics, such as Equal Accuracy and Equalized Odds, underscore the subjective nature of fairness and its context-dependent interpretation. Despite the presence of seemingly objective criteria, public perceptions of fairness vary widely, influenced by individual values and the specific context of decision-making scenarios.

The findings suggest that a one-size-fits-all approach to fairness is impractical in the design of recommendation systems. Instead, there is a need for flexible and adaptive models that can dynamically adjust fairness criteria based on the context and user preferences. This flexibility is essential for accommodating the broad spectrum of fairness perceptions and ensuring that systems are not only technically fair but also perceived as fair by their users.

Moreover, the study highlights the importance of transparency and user engagement in the development of recommendation systems. Clear communication about how algorithms work and the trade-offs involved in different fairness metrics is crucial for building trust and understanding among users. Engaging users in the design process through participatory methods can help align system functionalities with the diverse values and expectations of the user base, enhancing both fairness and acceptance.

One of the primary limitations of this study is the reliance on self-reported data from a relatively small and potentially non-representative sample of participants. The survey's limited reach, primarily targeting individuals connected to the higher education sector, may not fully capture the diversity of opinions and experiences found in the broader population. Additionally, the scenarios presented in the survey, while designed to represent high-stake and low-stake situations, may not encompass the full range of complexities and nuances encountered in real-world healthcare settings. The predefined fairness metrics provided to participants may also constrain their understanding and responses, as they do not account for all possible interpretations of fairness. Furthermore, the cross-sectional nature of the study means it captures participants' perceptions at a single point in time, without considering how these perceptions might evolve with increased knowledge or changing societal values. These limitations suggest that the findings should be interpreted with caution and underscore the need for further research to validate and expand upon these initial insights.

In conclusion, the pursuit of fairness in healthcare recommendation systems requires a nuanced approach that acknowledges the complexity of fairness perceptions. By considering the varied perspectives of users and the specificities of different decision-making contexts, developers can create more equitable and trusted systems. Future research should continue exploring the balance between fairness and other essential metrics, such as accuracy, and investigate methods for effectively communicating these complexities to the public. This ongoing effort is vital for the responsible and ethical deployment of AI-driven recommender systems in healthcare and beyond.

\bibliographystyle{ACM-Reference-Format}
\bibliography{references}


\begin{thebibliography}{16}


\ifx \showCODEN    \undefined \def \showCODEN     #1{\unskip}     \fi
\ifx \showDOI      \undefined \def \showDOI       #1{#1}\fi
\ifx \showISBNx    \undefined \def \showISBNx     #1{\unskip}     \fi
\ifx \showISBNxiii \undefined \def \showISBNxiii  #1{\unskip}     \fi
\ifx \showISSN     \undefined \def \showISSN      #1{\unskip}     \fi
\ifx \showLCCN     \undefined \def \showLCCN      #1{\unskip}     \fi
\ifx \shownote     \undefined \def \shownote      #1{#1}          \fi
\ifx \showarticletitle \undefined \def \showarticletitle #1{#1}   \fi
\ifx \showURL      \undefined \def \showURL       {\relax}        \fi
\providecommand\bibfield[2]{#2}
\providecommand\bibinfo[2]{#2}
\providecommand\natexlab[1]{#1}
\providecommand\showeprint[2][]{arXiv:#2}

\bibitem[Abu~Elyounes(2019)]%
        {abu_elyounes_contextual_2019}
\bibfield{author}{\bibinfo{person}{Doaa Abu~Elyounes}.} \bibinfo{year}{2019}\natexlab{}.
\newblock \bibinfo{title}{Contextual {Fairness}: {A} {Legal} and {Policy} {Analysis} of {Algorithmic} {Fairness}}.
\newblock
\newblock
\urldef\tempurl%
\url{https://doi.org/10.2139/ssrn.3478296}
\showDOI{\tempurl}


\bibitem[Elahi et~al\mbox{.}(2021)]%
        {elahi_beyond_2021}
\bibfield{author}{\bibinfo{person}{Mehdi Elahi}, \bibinfo{person}{Himan Abdollahpouri}, \bibinfo{person}{Masoud Mansoury}, {and} \bibinfo{person}{Helma Torkamaan}.} \bibinfo{year}{2021}\natexlab{}.
\newblock \showarticletitle{Beyond {Algorithmic} {Fairness} in {Recommender} {Systems}}. In \bibinfo{booktitle}{\emph{Adjunct {Proceedings} of the 29th {ACM} {Conference} on {User} {Modeling}, {Adaptation} and {Personalization}}} \emph{(\bibinfo{series}{{UMAP} '21})}. \bibinfo{publisher}{Association for Computing Machinery}, \bibinfo{address}{New York, NY, USA}, \bibinfo{pages}{41--46}.
\newblock
\showISBNx{978-1-4503-8367-7}
\urldef\tempurl%
\url{https://doi.org/10.1145/3450614.3461685}
\showDOI{\tempurl}


\bibitem[Hauptmann et~al\mbox{.}(2022)]%
        {hauptmann_research_2022}
\bibfield{author}{\bibinfo{person}{Hanna Hauptmann}, \bibinfo{person}{Alan Said}, {and} \bibinfo{person}{Christoph Trattner}.} \bibinfo{year}{2022}\natexlab{}.
\newblock \showarticletitle{Research directions in recommender systems for health and well-being}.
\newblock \bibinfo{journal}{\emph{User Modeling and User-Adapted Interaction}} \bibinfo{volume}{32}, \bibinfo{number}{5} (\bibinfo{date}{Nov.} \bibinfo{year}{2022}), \bibinfo{pages}{781--786}.
\newblock
\showISSN{1573-1391}
\urldef\tempurl%
\url{https://doi.org/10.1007/s11257-022-09349-4}
\showDOI{\tempurl}


\bibitem[Hiller(2020)]%
        {hiller_fairness_2020}
\bibfield{author}{\bibinfo{person}{Janine~S. Hiller}.} \bibinfo{year}{2020}\natexlab{}.
\newblock \showarticletitle{Fairness in the {Eyes} of the {Beholder}: {AI}; {Fairness}; and {Alternative} {Credit} {Scoring}}.
\newblock \bibinfo{journal}{\emph{West Virginia Law Review}} \bibinfo{volume}{123}, \bibinfo{number}{3} (\bibinfo{year}{2020}), \bibinfo{pages}{907--936}.
\newblock
\urldef\tempurl%
\url{https://heinonline.org/HOL/P?h=hein.journals/wvb123&i=937}
\showURL{%
\tempurl}


\bibitem[Köchling et~al\mbox{.}(2021)]%
        {kochling_highly_2021}
\bibfield{author}{\bibinfo{person}{Alina Köchling}, \bibinfo{person}{Shirin Riazy}, \bibinfo{person}{Marius~Claus Wehner}, {and} \bibinfo{person}{Katharina Simbeck}.} \bibinfo{year}{2021}\natexlab{}.
\newblock \showarticletitle{Highly {Accurate}, {But} {Still} {Discriminatory}}.
\newblock \bibinfo{journal}{\emph{Business \& Information Systems Engineering}} \bibinfo{volume}{63}, \bibinfo{number}{1} (\bibinfo{date}{Feb.} \bibinfo{year}{2021}), \bibinfo{pages}{39--54}.
\newblock
\showISSN{1867-0202}
\urldef\tempurl%
\url{https://doi.org/10.1007/s12599-020-00673-w}
\showDOI{\tempurl}


\bibitem[Li et~al\mbox{.}(2019)]%
        {li2019}
\bibfield{author}{\bibinfo{person}{Meng Li}, \bibinfo{person}{Helen~A. Colby}, {and} \bibinfo{person}{Philip Fernbach}.} \bibinfo{year}{2019}\natexlab{}.
\newblock \showarticletitle{Efficiency for Lives, Equality for Everything Else: How Allocation Preference Shifts Across Domains}.
\newblock \bibinfo{journal}{\emph{Social Psychological and Personality Science}} \bibinfo{volume}{10}, \bibinfo{number}{5} (\bibinfo{year}{2019}), \bibinfo{pages}{697--707}.
\newblock


\bibitem[Li et~al\mbox{.}(2021)]%
        {li_tutorial_2021}
\bibfield{author}{\bibinfo{person}{Yunqi Li}, \bibinfo{person}{Yingqiang Ge}, {and} \bibinfo{person}{Yongfeng Zhang}.} \bibinfo{year}{2021}\natexlab{}.
\newblock \showarticletitle{Tutorial on {Fairness} of {Machine} {Learning} in {Recommender} {Systems}}. In \bibinfo{booktitle}{\emph{Proceedings of the 44th {International} {ACM} {SIGIR} {Conference} on {Research} and {Development} in {Information} {Retrieval}}} \emph{(\bibinfo{series}{{SIGIR} '21})}. \bibinfo{publisher}{Association for Computing Machinery}, \bibinfo{address}{New York, NY, USA}, \bibinfo{pages}{2654--2657}.
\newblock
\showISBNx{978-1-4503-8037-9}
\urldef\tempurl%
\url{https://doi.org/10.1145/3404835.3462814}
\showDOI{\tempurl}


\bibitem[Rajkomar et~al\mbox{.}(2018)]%
        {rajkomar_ensuring_2018}
\bibfield{author}{\bibinfo{person}{Alvin Rajkomar}, \bibinfo{person}{Michaela Hardt}, \bibinfo{person}{Michael~D. Howell}, \bibinfo{person}{Greg Corrado}, {and} \bibinfo{person}{Marshall~H. Chin}.} \bibinfo{year}{2018}\natexlab{}.
\newblock \showarticletitle{Ensuring {Fairness} in {Machine} {Learning} to {Advance} {Health} {Equity}}.
\newblock \bibinfo{journal}{\emph{Annals of Internal Medicine}} \bibinfo{volume}{169}, \bibinfo{number}{12} (\bibinfo{date}{Dec.} \bibinfo{year}{2018}), \bibinfo{pages}{866--872}.
\newblock
\showISSN{0003-4819}
\urldef\tempurl%
\url{https://doi.org/10.7326/M18-1990}
\showDOI{\tempurl}
\newblock
\shownote{Publisher: American College of Physicians}.


\bibitem[Samuel(2022)]%
        {samuel_why_2022}
\bibfield{author}{\bibinfo{person}{Sigal Samuel}.} \bibinfo{year}{2022}\natexlab{}.
\newblock \bibinfo{title}{Why it’s so damn hard to make {AI} fair and unbiased}.
\newblock
\newblock
\urldef\tempurl%
\url{https://www.vox.com/future-perfect/22916602/ai-bias-fairness-tradeoffs-artificial-intelligence}
\showURL{%
\tempurl}


\bibitem[Saxena et~al\mbox{.}(2019)]%
        {saxena_how_2019}
\bibfield{author}{\bibinfo{person}{Nripsuta~Ani Saxena}, \bibinfo{person}{Karen Huang}, \bibinfo{person}{Evan DeFilippis}, \bibinfo{person}{Goran Radanovic}, \bibinfo{person}{David~C. Parkes}, {and} \bibinfo{person}{Yang Liu}.} \bibinfo{year}{2019}\natexlab{}.
\newblock \showarticletitle{How {Do} {Fairness} {Definitions} {Fare}? {Examining} {Public} {Attitudes} {Towards} {Algorithmic} {Definitions} of {Fairness}}. In \bibinfo{booktitle}{\emph{Proceedings of the 2019 {AAAI}/{ACM} {Conference} on {AI}, {Ethics}, and {Society}}} \emph{(\bibinfo{series}{{AIES} '19})}. \bibinfo{publisher}{Association for Computing Machinery}, \bibinfo{address}{New York, NY, USA}, \bibinfo{pages}{99--106}.
\newblock
\showISBNx{978-1-4503-6324-2}
\urldef\tempurl%
\url{https://doi.org/10.1145/3306618.3314248}
\showDOI{\tempurl}


\bibitem[Schlicker et~al\mbox{.}(2021)]%
        {schlicker_what_2021}
\bibfield{author}{\bibinfo{person}{Nadine Schlicker}, \bibinfo{person}{Markus Langer}, \bibinfo{person}{Sonja~K. Ötting}, \bibinfo{person}{Kevin Baum}, \bibinfo{person}{Cornelius~J. König}, {and} \bibinfo{person}{Dieter Wallach}.} \bibinfo{year}{2021}\natexlab{}.
\newblock \showarticletitle{What to expect from opening up ‘black boxes’? {Comparing} perceptions of justice between human and automated agents}.
\newblock \bibinfo{journal}{\emph{Computers in Human Behavior}}  \bibinfo{volume}{122} (\bibinfo{year}{2021}), \bibinfo{pages}{106837}.
\newblock
\showISSN{0747-5632}
\urldef\tempurl%
\url{https://doi.org/10.1016/j.chb.2021.106837}
\showDOI{\tempurl}


\bibitem[Schäfer et~al\mbox{.}(2017)]%
        {schafer_towards_2017}
\bibfield{author}{\bibinfo{person}{Hanna Schäfer}, \bibinfo{person}{Santiago Hors-Fraile}, \bibinfo{person}{Raghav~Pavan Karumur}, \bibinfo{person}{André Calero~Valdez}, \bibinfo{person}{Alan Said}, \bibinfo{person}{Helma Torkamaan}, \bibinfo{person}{Tom Ulmer}, {and} \bibinfo{person}{Christoph Trattner}.} \bibinfo{year}{2017}\natexlab{}.
\newblock \showarticletitle{Towards {Health} ({Aware}) {Recommender} {Systems}}. In \bibinfo{booktitle}{\emph{Proceedings of the 2017 {International} {Conference} on {Digital} {Health}}} \emph{(\bibinfo{series}{{DH} '17})}. \bibinfo{publisher}{Association for Computing Machinery}, \bibinfo{address}{New York, NY, USA}, \bibinfo{pages}{157--161}.
\newblock
\showISBNx{978-1-4503-5249-9}
\urldef\tempurl%
\url{https://doi.org/10.1145/3079452.3079499}
\showDOI{\tempurl}


\bibitem[Srivastava et~al\mbox{.}(2019)]%
        {srivastava_mathematical_2019}
\bibfield{author}{\bibinfo{person}{Megha Srivastava}, \bibinfo{person}{Hoda Heidari}, {and} \bibinfo{person}{Andreas Krause}.} \bibinfo{year}{2019}\natexlab{}.
\newblock \showarticletitle{Mathematical {Notions} vs. {Human} {Perception} of {Fairness}: {A} {Descriptive} {Approach} to {Fairness} for {Machine} {Learning}}. In \bibinfo{booktitle}{\emph{Proceedings of the 25th {ACM} {SIGKDD} {International} {Conference} on {Knowledge} {Discovery} \& {Data} {Mining}}} \emph{(\bibinfo{series}{{KDD} '19})}. \bibinfo{publisher}{Association for Computing Machinery}, \bibinfo{address}{New York, NY, USA}, \bibinfo{pages}{2459--2468}.
\newblock
\showISBNx{978-1-4503-6201-6}
\urldef\tempurl%
\url{https://doi.org/10.1145/3292500.3330664}
\showDOI{\tempurl}


\bibitem[Wang et~al\mbox{.}(2023)]%
        {wang_survey_2023}
\bibfield{author}{\bibinfo{person}{Yifan Wang}, \bibinfo{person}{Weizhi Ma}, \bibinfo{person}{Min Zhang}, \bibinfo{person}{Yiqun Liu}, {and} \bibinfo{person}{Shaoping Ma}.} \bibinfo{year}{2023}\natexlab{}.
\newblock \showarticletitle{A {Survey} on the {Fairness} of {Recommender} {Systems}}.
\newblock \bibinfo{journal}{\emph{ACM Trans. Inf. Syst.}} \bibinfo{volume}{41}, \bibinfo{number}{3} (\bibinfo{date}{Feb.} \bibinfo{year}{2023}), \bibinfo{pages}{52:1--52:43}.
\newblock
\showISSN{1046-8188}
\urldef\tempurl%
\url{https://doi.org/10.1145/3547333}
\showDOI{\tempurl}


\bibitem[Wu et~al\mbox{.}(2020)]%
        {wu_development_2020}
\bibfield{author}{\bibinfo{person}{Guangyao Wu}, \bibinfo{person}{Pei Yang}, \bibinfo{person}{Yuanliang Xie}, \bibinfo{person}{Henry~C. Woodruff}, \bibinfo{person}{Xiangang Rao}, \bibinfo{person}{Julien Guiot}, \bibinfo{person}{Anne-Noelle Frix}, \bibinfo{person}{Renaud Louis}, \bibinfo{person}{Michel Moutschen}, \bibinfo{person}{Jiawei Li}, \bibinfo{person}{Jing Li}, \bibinfo{person}{Chenggong Yan}, \bibinfo{person}{Dan Du}, \bibinfo{person}{Shengchao Zhao}, \bibinfo{person}{Yi Ding}, \bibinfo{person}{Bin Liu}, \bibinfo{person}{Wenwu Sun}, \bibinfo{person}{Fabrizio Albarello}, \bibinfo{person}{Alessandra D'Abramo}, \bibinfo{person}{Vincenzo Schininà}, \bibinfo{person}{Emanuele Nicastri}, \bibinfo{person}{Mariaelena Occhipinti}, \bibinfo{person}{Giovanni Barisione}, \bibinfo{person}{Emanuela Barisione}, \bibinfo{person}{Iva Halilaj}, \bibinfo{person}{Pierre Lovinfosse}, \bibinfo{person}{Xiang Wang}, \bibinfo{person}{Jianlin Wu}, {and} \bibinfo{person}{Philippe Lambin}.} \bibinfo{year}{2020}\natexlab{}.
\newblock \showarticletitle{Development of a clinical decision support system for severity risk prediction and triage of {COVID}-19 patients at hospital admission: an international multicentre study}.
\newblock \bibinfo{journal}{\emph{European Respiratory Journal}} \bibinfo{volume}{56}, \bibinfo{number}{2} (\bibinfo{date}{Aug.} \bibinfo{year}{2020}).
\newblock
\showISSN{0903-1936, 1399-3003}
\urldef\tempurl%
\url{https://doi.org/10.1183/13993003.01104-2020}
\showDOI{\tempurl}
\newblock
\shownote{Publisher: European Respiratory Society Section: Original Articles}.


\bibitem[Wu et~al\mbox{.}(2023)]%
        {wu_fairness_2023}
\bibfield{author}{\bibinfo{person}{Yao Wu}, \bibinfo{person}{Jian Cao}, {and} \bibinfo{person}{Guandong Xu}.} \bibinfo{year}{2023}\natexlab{}.
\newblock \showarticletitle{Fairness in {Recommender} {Systems}: {Evaluation} {Approaches} and {Assurance} {Strategies}}.
\newblock \bibinfo{journal}{\emph{ACM Trans. Knowl. Discov. Data}} \bibinfo{volume}{18}, \bibinfo{number}{1} (\bibinfo{date}{Aug.} \bibinfo{year}{2023}), \bibinfo{pages}{10:1--10:37}.
\newblock
\showISSN{1556-4681}
\urldef\tempurl%
\url{https://doi.org/10.1145/3604558}
\showDOI{\tempurl}


\end{thebibliography}

\end{document}